\def\year{2019}\relax
\documentclass[letterpaper]{article} 
\usepackage{aaai19}  
\usepackage{times}  
\usepackage{helvet}  
\usepackage{courier}  
\usepackage{url}  
\usepackage{graphicx}  
\usepackage{amsmath}
\usepackage{amssymb}
\usepackage{pifont}
\usepackage[ruled,noend]{algorithm2e}
\usepackage{wrapfig,lipsum,booktabs}
\usepackage{comment}

\usepackage{footnote}
\makesavenoteenv{table}
\usepackage{scrextend}
\usepackage{multirow}
\usepackage{tabularx}
\usepackage{subcaption,booktabs}
\DeclareMathOperator*{\argmin}{arg\,min}  

\title{ A New Ensemble Learning Framework for 3D Biomedical Image Segmentation }
\author{ 
Hao Zheng\thanks{Equal contribution (Co-first authors). Author ordering determined by dice rolling over a Google Hangout.},  
Yizhe Zhang\footnotemark[1], 
Lin Yang\footnotemark[1], \\
{\bf \Large
Peixian Liang, Zhuo Zhao, Chaoli Wang, Danny Z. Chen}\\
Department of Computer Science and Engineering,
University of Notre Dame, Notre Dame, IN 46556, USA \\
\{hzheng3, yzhang29, lyang5, pliang, zzhao3, cwang11, dchen\}@nd.edu
} 
\begin{document}
\maketitle

\begin{abstract}
3D image segmentation plays an important role in biomedical image analysis. Many 2D and 3D deep learning models have achieved state-of-the-art segmentation performance on 3D biomedical image datasets. Yet, 2D and 3D models have their own strengths and weaknesses, and by unifying them together, one may be able to achieve more accurate results. In this paper, we propose a new {\em ensemble learning} framework for 3D biomedical image segmentation that combines the merits of 2D and 3D models. First, we develop a fully convolutional network based meta-learner to learn how to improve the results from 2D and 3D models (base-learners). Then, to minimize over-fitting for our sophisticated meta-learner, we devise a new training method that uses the results of the base-learners as multiple versions of ``ground truths''. Furthermore, since our new meta-learner training scheme does not depend on manual annotation, it can utilize abundant unlabeled 3D image data to further improve the model. Extensive experiments on two public datasets (the HVSMR 2016 Challenge dataset and the mouse piriform cortex dataset) show that our approach is effective under fully-supervised, semi-supervised, and transductive settings, and attains superior performance over state-of-the-art image segmentation methods.
\end{abstract}

\section{Introduction}\label{Sec-Intro}
3D image segmentation plays an important role in biomedical image analysis (e.g., segmenting the whole heart to diagnose cardiac diseases \cite{pace2015interactive,yu2017automatic} and segmenting neuronal structures to identify cortical connectivity \cite{lee2015recursive,neuron2}). With recent rapid advances in deep learning, many 2D \cite{neuron2,wolterink2016dilated} and 3D \cite{yu2017automatic,cciccek20163d,chen2017voxresnet} convolutional neural networks (CNNs) have been developed to attain state-of-the-art segmentation results on various 3D biomedical image datasets \cite{pace2015interactive,neuron2}.
However, due to the limitations of both GPU memory and computing power, when designing 2D/3D CNNs for 3D biomedical image segmentation, the trade-off between the field of view and utilization of inter-slice information in 3D images remains a major concern.
For example, 3D CNNs attempt to fully utilize 3D image information but only have a limited field of view (e.g., $64\times64\times64$ \cite{yu2017automatic}), while 2D CNNs can have a much larger field of view (e.g., $572\times572$ \cite{ronneberger2015u}) but are not able to fully explore inter-slice information.

Many methods have been proposed to circumvent this trade-off by carefully designing the structures of 2D and 3D CNNs.
Their main ideas can be broadly classified into two categories. 
The models in the first category selectively choose the input data. For example, the tri-planar schemes \cite{wolterink2016dilated,prasoon2013deep} use only three orthogonal planes (i.e., the $xy$, $yz$, and $xz$ planes) instead of the whole 3D image, aiming to utilize inter-slice information without sacrificing the field of view. 
The models in the second category first summarize intra-slice information using 2D CNNs and then use the distilled information as an (extra) input to their 3D network component. For example, in \cite{lee2015recursive}, intra-slice information is first extracted using a 2D CNN (VD2D) and then passed to the 3D component (VD2D3D) via recursive training. In \cite{chen2016combining}, its recurrent neural network (RNN) component directly uses the results of 2D CNNs as input to compute 3D segmentation.

However, these methods still have considerable drawbacks. Tri-planar schemes \cite{wolterink2016dilated,prasoon2013deep} use only a small fraction of 3D image information and the computation cost is not reduced but shifted to the inference stage (a tri-planar scheme can only predict one voxel at a time, which is very slow when predicting new 3D images). For models that first summarize intra-slice information, the asymmetry nature of the network design (first 2D, then 3D) may hinder a full utilization of 3D image information (2D results may dominate since they are much easier to be interpreted than raw images in the 3D stage).

\begin{figure*}
  \centering
  \includegraphics[width=0.7\textwidth]{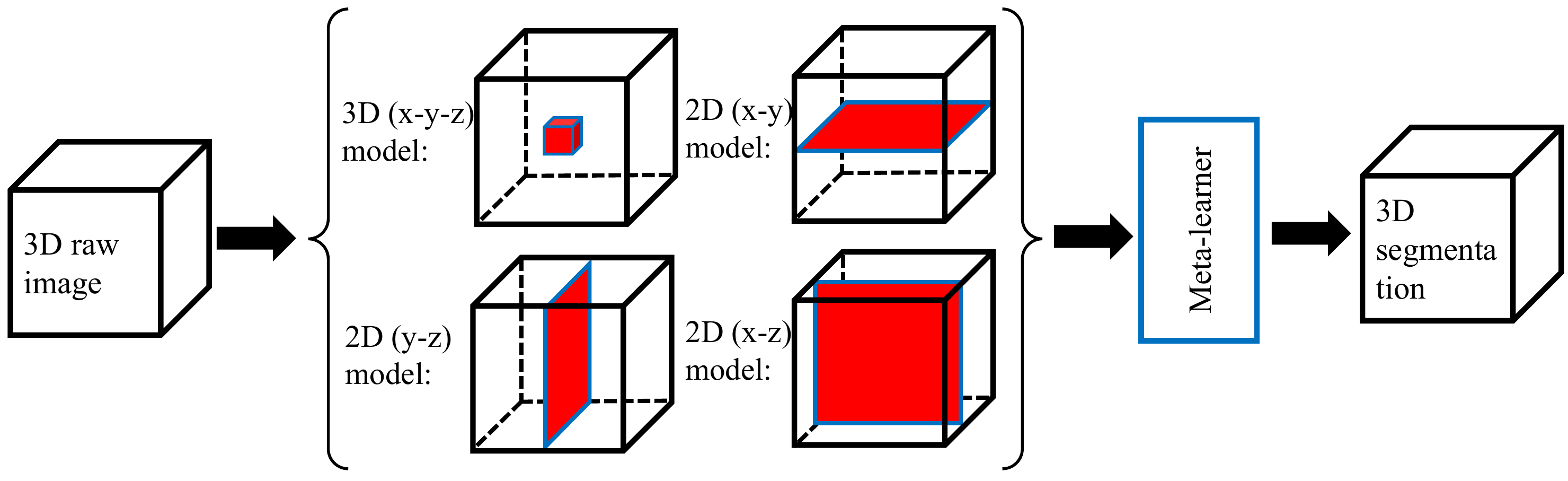}
  \caption{An overview of our proposed framework. Red box/planes show the effective fields of view of the corresponding 3D/2D base-learners. Our meta-learner works on top of all the base-learners.}
\label{fig:overview}
\end{figure*}

In this paper, we explore a different perspective. Instead of designing new network structures to circumvent the trade-off between field of view and inter-slice information, we address this difficulty by developing a new {\em ensemble learning} framework for 3D biomedical image segmentation which aims to retain and combine the merits of 2D/3D models. Fig.~\ref{fig:overview} gives an overview of our framework.

Due to the heterogeneous nature of our 2D and 3D models (base-learners), we use the idea of stacking \cite{wolpert1992stacked} (i.e., training a meta-learner to combine the results of multiple base-learners).
Given a set of image samples, $X$ = $\{x_1, x_2, \ldots, x_n\}$, their corresponding ground truth, $Y$ = $\{y_1,y_2, \ldots, y_n\}$, and a set of base-learners, $F$ = $\{f_1, f_2, \ldots, f_m\}$, a common design of a meta-learner is to learn the prediction of $(x_i, \hat{y_i})$ = $f_{meta}(f_1(x_i),f_2(x_i), \ldots, f_m(x_i))$, by fitting $y_i$. Since the results from the base-learners can be very close to the ground truth, training the meta-learner by directly fitting $y_i$ is susceptible to over-fitting.
Many measures have been proposed to address this issue: (1) simple meta-learner structure designs (e.g., in \cite{ju2018relative}, the meta-learner was implemented as a single $1 \times 1$ convolution layer, and in \cite{makarchuk2018ensembling}, the meta-learner was implemented using the XGBoost classifier \cite{chen2016xgboost}); (2) excluding raw image information from the input \cite{zhou2012ensemble}; (3) splitting the training data into multiple folds and training the meta-learner by minimizing the cross-validated risk \cite{van2007super}.

However, in our 3D biomedical image segmentation scenario, these meta-learner designs may not work well due to the following reasons. First, each of our individual base-learners (2D and 3D models) has its distinct merit; in many difficult image areas, it is quite likely that only one of the base-learners could produce the correct results. Thus, our meta-learner should be sophisticated enough in order to capture the merits of all the base-learners. Second, since extensive annotation efforts are often needed to produce full 3D annotation, not many 3D training images are available (e.g., 3 in \cite{lee2015recursive} and 10 in \cite{pace2015interactive}) in common 3D biomedical image datasets. Splitting the already scarce training data can largely lower the accuracy of the base-learners and meta-learner.

Hence, we propose a new stacking method that includes (1) a deep learning based meta-learner to combine and improve the results of the base-learners, and (2) a new meta-learner training method that can train a sophisticated meta-learner with less risk of over-fitting.

\textbf{A deep learning based meta-learner.}
Comparing with image classification, the output domain of image segmentation is much more structural. However, 
recent studies have not leveraged this property to design a better meta-learner for image segmentation. For example, in \cite{ju2018relative,nigam2018ensemble}, only linear combination of base-learners was explored.
We develop a new fully convolutional network (FCN) based meta-learner to capture the merits of our base-learners and produce spatially consistent results.

\textbf{Minimizing the risk of over-fitting.}
A key idea of our meta-learner training method is to use the results of the base-learners as {\it pseudo-labels} \cite{lee2013pseudo} and compute ensemble by finding a balance among these pseudo-labels through the training process (in contrast to directly fitting $y_i$). More specifically, for each input sample $x_i$, there are multiple versions of pseudo-labels (from the individual base-learners). During the iterative meta-learner training process, in each iteration, we randomly choose one pseudo-label from all the versions and use it as ``ground truth'' to compute the loss and update meta-learner parameters.
Iteration by iteration, the pseudo-labels with small disagreement would provide a more consistent supervision and the pseudo-labels with large disagreement would request the meta-learner to find a balanced solution by minimizing the overall loss.

Our method can minimize the risk of over-fitting in two aspects. (1) Intuitively, over-fitting occurs when a model over-explains the ground truth. Since our method uses multiple versions of ``ground truths'' (pseudo-labels), the meta-learner is unlikely to over-fit any one of them. (2) Since our meta-learner training uses only model-generated results, unlabeled data can be easily incorporated into the training process; this will allow us to further reduce over-fitting.

Compared with previous methods that combine 2D and 3D models, our main contributions are:
(a) a new ensemble learning framework for tackling 3D biomedical image segmentation from a different perspective, and
(b) an effective meta-learner training method for ensemble learning that minimizes the risk of over-fitting and makes use of unlabeled data.
Extensive experiments on two public datasets (the HVSMR 2016 Challenge dataset \cite{pace2015interactive} and the mouse piriform cortex dataset \cite{lee2015recursive}) show that our framework is effective under fully-supervised, semi-supervised, and transductive settings, and attains superior performance over the state-of-the-art methods \cite{yu2017automatic,neuron2,chen2017voxresnet}.
Code will be made publicly available at
\mbox{\url{https://github.com/HaoZheng94/Ensemble}}.

\section{Method}\label{Sec-Method}
Our proposed approach has two main components:
(1) a group of 2D and 3D base-learners that are trained to explore the training data from different geometric perspectives;
(2) an ensemble learning framework that uses a deep learning based meta-learner to combine the results from the base-learners. A schematic overview of our proposed framework is shown in Fig.~\ref{fig:overview}.

In Section \ref{sec:constructBL}, we illustrate how to design our 2D and 3D base-learners to achieve a set of accurate and diverse results. In Section \ref{sec:meta-learner}, we discuss our new deep learning based meta-learner that can considerably improve the results from the base-learners. In Section \ref{sec:trainingMethod}, we present our new method for training a more powerful meta-learner while preventing over-fitting.

\begin{figure*}[ht]
  \centering
  \includegraphics[width=0.76\textwidth]{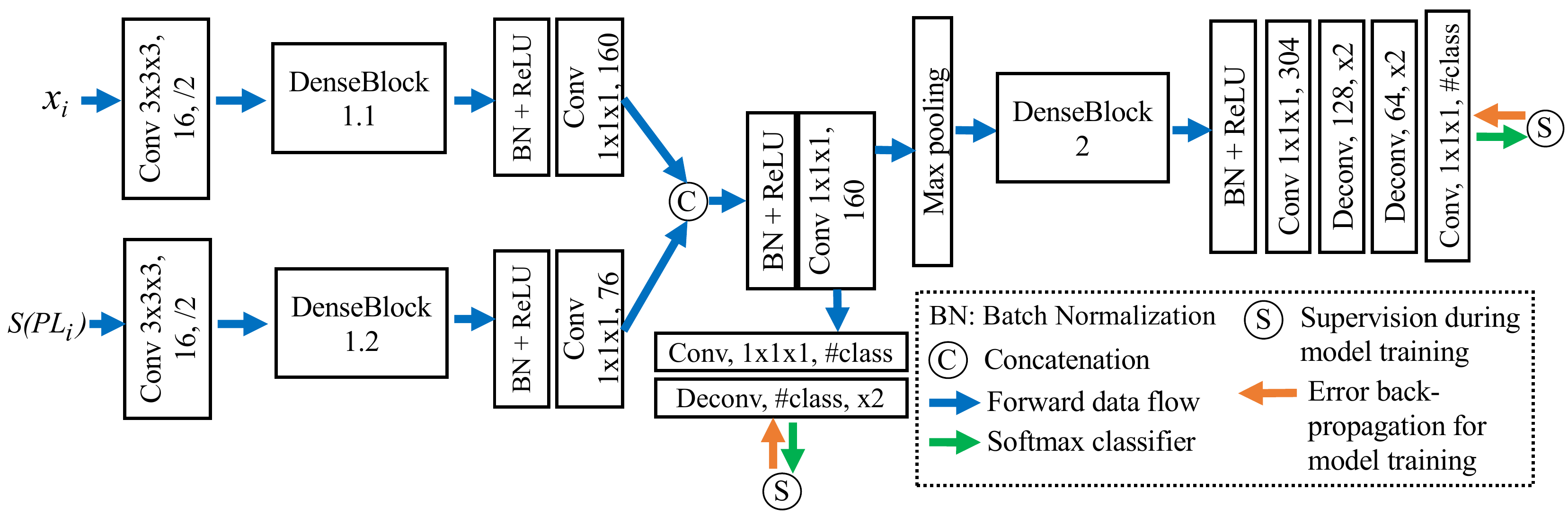}
  \caption{Our deep meta-learner (a variant of 3D DenseVoxNet \cite{yu2017automatic}). Since $S(PL_i)$ and $x_i$ are of different nature, we use separate encoding blocks (i.e., DenseBlock 1.1 and DenseBlock 1.2) for extracting information from $S(PL_i)$ and $x_i$, respectively, before the information fusion. The auxiliary loss in the side path can improve the gradient flow within the network.}
\label{fig:dml}
\end{figure*}

\subsection{2D and 3D base-learners}\label{sec:constructBL}
To achieve the best possible ensemble results, it is commonly desired that individual base-learners be as {\em accurate} and as {\em diverse} as possible \cite{zhou2012ensemble}. In this section, we show how to design our 2D and 3D base-learners to satisfy these two criteria.

\textbf{For accurate results.} Our 2D model basically follows the structure of that in \cite{yang2017suggestive}. We choose this structure because it is based on a well-known FCN \cite{chen2016dcan} which has attained lots of successes in biomedical image segmentation and has been integrated in recent advances of deep learning network design structures, such as batch normalization \cite{ioffe2015batch}, residual networks and bottleneck design \cite{he2016deep}. It generalizes well and is fast to train. As for the 3D model, we use DenseVoxNet \cite{yu2017automatic}, for three reasons.  First, it adopts a 3D FCN architecture, and thus can fully incorporate 3D image cues and geometric cues for effective volume-to-volume prediction. Second, it utilizes the state-of-the-art dense connectivity \cite{huang2017densely} to accelerate the training process, improve parameters and computational efficiency, and maintain abundant (both low- and high-complexity) features for segmenting complicated biomedical structures. Third, it takes advantage of auxiliary side paths for deep supervision \cite{dou20163d} to improve the gradient flow within the network and stabilize the learning process. For further details of these 2D and 3D models, the readers are referred to \cite{yang2017suggestive} and \cite{yu2017automatic}.

\textbf{For diverse results.} Our key idea to achieve diverse results is to let each of our base-learners have a unique geometric view of the data. As discussed in Section~\ref{Sec-Intro}, 
2D models can have large fields of view in 2D slices while 3D models can better utilize 3D image information in a smaller field of view. Our mix of 2D and 3D base-learners creates the first level of diversity. To further boost diversity, within the group of 2D base-learners, we leverage the 3D image data to create multiple 2D views (representations) of the 3D images (e.g., $xy$, $xz$, and $yz$ views). The different 2D representations of the 3D images create 2D models with diverse strengths (e.g., large fields of view for different planes) and thus generate diverse 2D results. Note that the 2D representations are not limited to being orthogonal to each of the major axes. But, based on our trial studies, we found that the results from the $xy$, $xz$, and $yz$ views are the most accurate (probably because no interpolation is needed when extracting these 2D slices from 3D images) and already create good diversity. 

Thus, in our framework, we use the following four base-learners: a 3D DenseVoxNet \cite{yu2017automatic} for utilizing full 3D information; three 2D FCNs \cite{yang2017suggestive} for large fields of view in the $xy$, $xz$, and $yz$ planes.

\subsection{Deep meta-learner structure design}\label{sec:meta-learner}
Since our base-learners have distinct model architectures and work on different geometric perspectives of 3D images to produce diverse predictions, for difficult image areas, there is a better chance that one of the base-learners would give correct predictions (see Fig.~\ref{fig:Comparison}). In order to attain a meta-learner to pick up the correct predictions, we need a model that is, architecture and complexity wise, capable of learning robust visual features for jointly utilizing the diverse prediction results (from the base-learners) as well as the raw image information. It is known that simple models (e.g., linear models, shallow neural networks) are not powerful enough to learn/extract robust and comprehensive features for difficult vision tasks \cite{zhou2012ensemble}. Furthermore, our learning task is a segmentation problem that requires spatially consistent output. Thus, we employ a state-of-the-art 3D FCN (DenseVoxNet \cite{yu2017automatic}) for building our meta-learner. The input of the network is the base-learners' results and the raw image, and the output of the network is the computed ensemble. Below we describe how to construct the input of our deep meta-learner and the details of the deep meta-learner's model architecture.

Given a set of image samples, $X=\{x_1, x_2, \ldots, x_n\}$, and a set of base-learners, $F$ $=$ $\{f_1$, $f_2$, $\ldots, f_m\}$, a {\it pseudo-label set} for each $x_i$ can be obtained as $PL_i=\{f_1(x_i),f_2(x_i), \ldots, f_m(x_i)\}$. The input of our meta-learner $\mathcal{H}$ includes two parts: $x_i$ and $S(PL_i)$, where $S$ is a function of $PL_i$ that forms a representation of $PL_i$. There are multiple design choices for constructing $S$. For example, (1) concatenating all the elements of $PL_i$, or (2) averaging all the elements of $PL_i$. Concatenation allows the meta-learner to gain full information from the base-learners (no information is added or lost). Averaging provides a more compact representation of all pseudo-labels, while still showing the image areas where the pseudo-labels hold agreement or disagreement. Furthermore, using the average of all the pseudo-labels of $x_i$ to form part of the meta-learner's input can be viewed as a preliminary ensemble of the base-learners. We have experimented with both these design choices and found that making $S$ an averaging function of the elements of $PL_i$ gives slightly better results.  
The overall model specification of our proposed deep meta-learner is shown in Fig.~\ref{fig:dml}.

\subsection{Meta-learner training using pseudo-labels}\label{sec:trainingMethod}
A major goal of our training procedure is to train a powerful meta-learner, while minimizing the risk of over-fitting. To achieve this goal, instead of using the ground truth to supervise the meta-learner training, we use the pseudo-labels produced by our base-learners (as discussed in Section~\ref{sec:constructBL}) %
to form the supervision signals. Because there are multiple possible targets (pseudo-labels) for the meta-learner to fit, the meta-learner is unlikely to over-fit any fixed target. The base-learners can also be applied to generate pseudo-labels for unlabeled data. Thus, our method is also capable of using unlabeled data for deep meta-learner training (which can further reduce over-fitting).

Suppose $X$, $PL_{i}$, and $S(PL_{i})$ are given, for $i=1,2, \ldots, n$. Ideally, the learning objective of the meta-learner would be: (1) finding the ``best'' pseudo-labels in $PL_i$, and (2) training the meta-learner $\mathcal{H}$ to fit the pseudo-labels found in (1). However, the best pseudo-labels are not clearly defined and can be difficult to find. Based on different evaluation criteria, the ``best'' choices can be different. Even when a criterion is given, using the most accurate pseudo-labels can likely lead to a higher chance of suffering over-fitting. Furthermore, when training using unlabeled data, it is in general quite difficult to determine which pseudo-label gives more accurate predictions than the others. One could set up a hand-crafted algorithm based on a predefined criterion to select the ``best'' pseudo-labels for training. The meta-learner, however, could very likely over-fit the algorithm's choices and hence likely not be able to generalize well to future unseen image data.

Rather than explicitly defining the full learning objective for meta-learner training, we initially train the meta-learner in order to set up a near-optimal (or sub-optimal) configuration: The meta-learner is aware of all the available pseudo-labels, and its position in the hypothesis space is influenced by the raw image and the pseudo-label data distribution. Next, the meta-learner itself chooses the nearest pseudo-labels to fit (based on its current model parameters) and updates its model parameters based on its current choices. This nearest-neighbor-fit process iterates until the meta-learner fits the nearest neighbors well enough. Thus, our meta-learner training consists of two phases: (1) random-fit, and (2) nearest-neighbor-fit. We describe these two training phases below.

\begin{algorithm}[t]
    \SetAlgoLined
    \KwIn{$(x_i,PL_i=\{f_1(x_i),f_2(x_i), \ldots, f_m(x_i)\},S(PL_i))$ , $i=1,2, \ldots, n$\\
    \KwOut{A trained meta-learner $\mathcal{H}$}
    initialize a meta-learner $\mathcal{H}$ with random weights;
    mini-batch = $\emptyset$\\
    \While{stopping condition not met}{
        \For{$k$ = 1 to batch-size}{
            $p=rand$-$int(1,n)$\\
            $q=rand$-$int(1,m)$\\
            add training sample $\{(x_p,S(PL_{p})),f_q(x_p)\}$ to the mini-batch\
        }
        update $\mathcal{H}$ using training samples in the mini-batch with forward and backward propagation\\
        mini-batch = $\emptyset$\
        }
    }
\caption{Random-fit}
\label{alg:randomfit}
\end{algorithm}

\textbf{Random-fit.} 
In the first training phase (which aims to train the meta-learner $\mathcal{H}$ to reach a near-optimal solution), we seek to minimize the overall cross-entropy loss for all the image samples with respect to all the pseudo-labels:
\begin{equation}\label{eq1}
\text{\footnotesize ${\ell(\theta_{\mathcal{H}})}{=}
{\sum_{i=1}^{n}}{\sum_{j=1}^{m}} {\ell_{mce}}(\theta_{\mathcal{H}}(x_i,S(PL_i)){,}f_j(x_i))$},
\end{equation}
where $\theta_{\mathcal{H}}$ is the meta-learner's model parameters and ${\ell_{mce}}$ is a multi-class cross-entropy criterion.  
The above loss ensures that the meta-learner training process in this phase works on all the available pseudo-labels. Since the loss function itself does not impose any favor towards any particular pseudo-labels produced by the base-learners, our meta-learner is unlikely to over-fit any pseudo-labels. Exploring the overall raw image and the pseudo-label data distribution, the meta-learner obtained by minimizing the above loss may have different tendencies towards different pseudo-labels.

To effectively optimize the loss function in Eq.~(\ref{eq1}), we develop a random-fit algorithm. In the SGD-based optimization, for one image sample $x_i$, our algorithm randomly chooses a pseudo-label from $PL_i$ and sets it as the current ``ground truth'' for $x_i$ (see Algorithm \ref{alg:randomfit}). This ensures the supervision signals not to impose any bias towards any base-learner, and allows image samples with diverse pseudo-labels to have a better chance to be influenced by other image samples.
Our experiments show that our random-fit algorithm is effective for learning with diverse pseudo-labels.

\begin{algorithm}[t]
    \SetAlgoLined
    \KwIn{$(x_i,PL_i=\{f_1(x_i),f_2(x_i), \ldots, f_m(x_i)\},S(PL_i))$ , $i=1,2, \ldots, n$, \newline
    meta-learner $\mathcal{H}$ (obtained from random-fit) \\
    \KwOut{A refined meta-learner $\mathcal{H}$}
    mini-batch = $\emptyset$\\
    \While{stopping condition not met}{
        \For{$k$ = 1 to batch-size}{
            $p=rand$-$int(1,n)$\\
            $\hat{y}=\mathcal{H}(x_p,S(PL_p))$\\
            $\hat{q}=\argmin_{q=1,2, \ldots, m}\mathcal{L}_{mce}(\hat{y},f_q(x_p))$
            
            add training sample $\{(x_p,S(PL_{p})),f_{\hat{q}}(x_p)\}$ to the mini-batch\
        }
        update $\mathcal{H}$ using training samples in the mini-batch with forward and backward propagation\\
        mini-batch = $\emptyset$
        }
    }
\caption{Nearest-neighbor-fit  (NN-fit)}
\label{alg:nnfit}
\end{algorithm}

\textbf{Nearest-neighbor-fit (NN-fit).} 
Unlike image classification problems, the label space of segmentation problems is with high spatial dimensions and not all solutions in the label space are meaningful. For example, a union or intersection of two prediction maps (pseudo-labels) may incur a risk of yielding strange shapes or structures that are quite likely incorrect. Even when all pseudo-labels of a particular image sample are close to the true solution, the trained meta-learner, if not fitting any of the pseudo-labels appropriately, can still have a risk of producing new types of errors.

\begin{figure}
  \centering
  \includegraphics[width=0.99\linewidth]{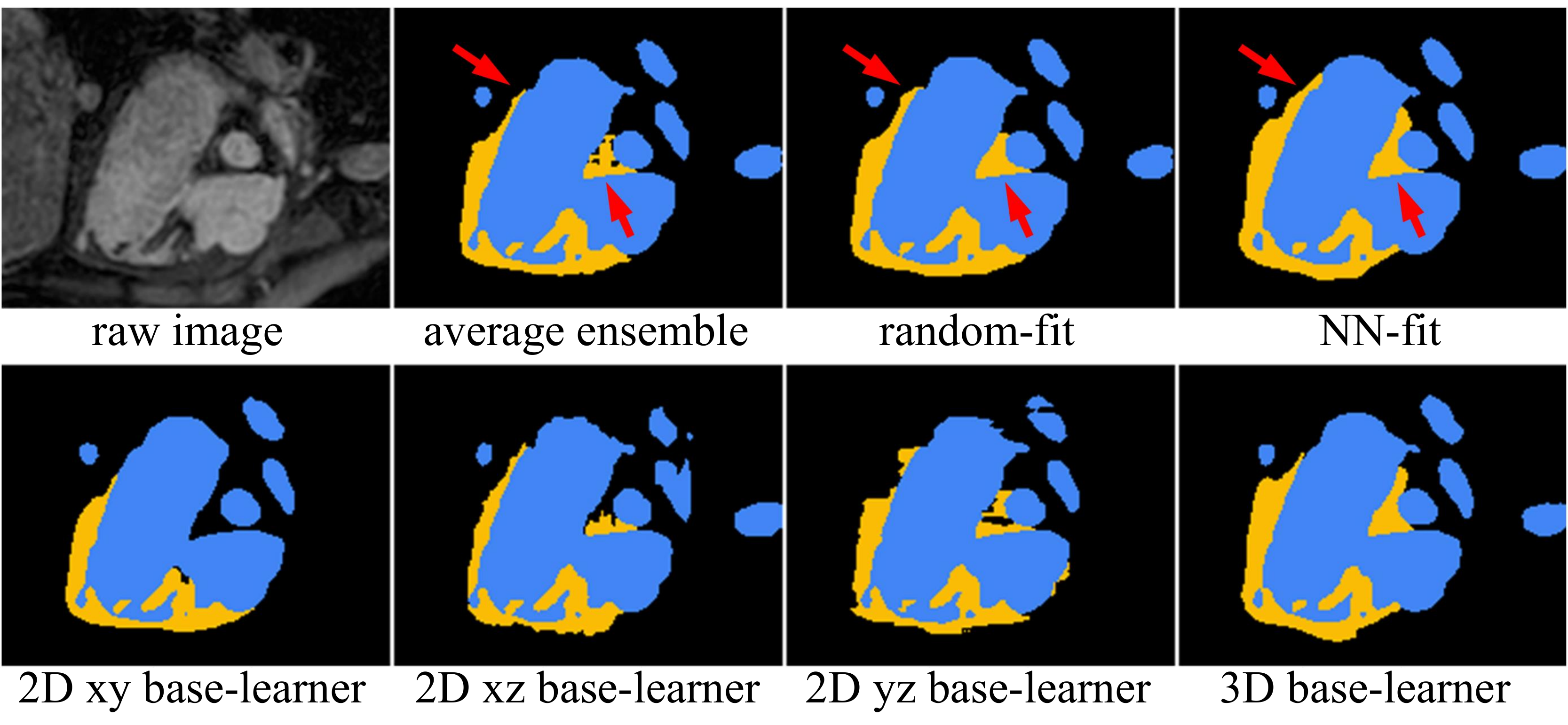}
  \caption{Visual comparison of segmentation results (yellow: myocardium; blue: blood pool). With NN-fit, our meta-learner can achieve more accurate segmentation of myocardium (red arrows).}
  \label{fig:Comparison}
\end{figure}

Thus, to help the model training process converge, in the second training phase, we aim to train the meta-learner to fit the nearest pseudo-label. Since the overall training loss is based on cross-entropy, to make NN-fit have direct effects on the convergence of the model training, we use cross-entropy to measure difference between a meta-learner's output and a pseudo-label. The details of our NN-fit algorithm are presented in Algorithm \ref{alg:nnfit}. 
Our experiments show that NN-fit can effectively improve the performance of the deep meta-learner (see Fig.~\ref{fig:Comparison}, and Tables~\ref{tab:HVSMR_1} and \ref{tab:Neuron}).

\begin{table*}
\centering
\caption{Quantitative analysis on the HVSMR 2016 dataset. VFN$^\ast$: For fair comparison, we use DenseVoxNet \cite{yu2017automatic} as backbone, which is the same as our 3D base-learner.
}
\label{tab:HVSMR_1}
\scriptsize
\begin{tabular}{c ccc ccc c}
    \hline
    \multirow{ 2}{*}{ Method } &   \multicolumn{3}{c }{Myocardium} & \multicolumn{3}{c }{Blood pool}  &  \multirow{ 2}{*}{\shortstack{Overall \\ score}} \\  \cline{2-7}
    & Dice   &  ADB [$mm$]    &  Hausdorff [$mm$]   & Dice   &  ADB [$mm$]    &  Hausdorff [$mm$] &  \\ 
    \hline
    3D U-Net \cite{cciccek20163d}   &  0.694 $\pm$ 0.076  &  1.461 $\pm$ 0.397  & 10.221 $\pm$ 4.339  &   0.926 $\pm$ 0.016  &  0.940 $\pm$ 0.192   &  8.628 $\pm$ 3.390    &   -0.419  \\ 
    
    VoxResNet \cite{chen2017voxresnet}  & 0.774 $\pm$ 0.067  &   1.026 $\pm$ 0.400   &  6.572 $\pm$ 0.013  &  0.929 $\pm$ 0.013  &  0.981 $\pm$ 0.186  &  9.966 $\pm$ 3.021     &  -0.202   \\ 
   
    DenseVoxNet  \cite{yu2017automatic} &  0.821 $\pm$ 0.041  &  0.964 $\pm$ 0.292  &  7.294 $\pm$ 3.340  & 0.931 $\pm$ 0.011  &  0.938 $\pm$ 0.224  &  9.533 $\pm$ 4.194    &   -0.161 \\ 
  
    Wolterink \emph{et al.}\cite{wolterink2016dilated}  &  0.802 $\pm$ 0.060   &  0.957 $\pm$ 0.302    &  6.126 $\pm$ 3.565  &  0.926 $\pm$ 0.018  & 0.885 $\pm$ 0.223  &  7.069 $\pm$ 2.857  &  -0.036   \\ 

    VFN$^\ast$~\cite{xia2018bridging}  &  0.773 $\pm$ 0.098 &  0.877 $\pm$ 0.318   &  4.626 $\pm$ 2.319   &   0.935 $\pm$ 0.009  & 0.770 $\pm$ 0.098 &   5.420 $\pm$ 2.152  &   0.108   \\ 
    \hline
    
    Base learner 2D ($xy$)  & 0.789 $\pm$ 0.076  &  0.852 $\pm$ 0.265  & 4.231 $\pm$ 1.908  &  0.930 $\pm$ 0.016 & 0.794 $\pm$ 0.153  &  5.295 $\pm$ 1.671 & 0.13   \\ 

    Base learner 2D ($xz$)  &  0.736 $\pm$ 0.093  &  1.000 $\pm$ 0.260   &  5.417 $\pm$ 1.604  &  0.924 $\pm$ 0.015  &  0.932 $\pm$ 0.113  &  7.951 $\pm$ 2.820 & -0.098 \\ 

    Base learner 2D ($yz$)  &  0.756 $\pm$ 0.082  &  0.870 $\pm$ 0.181   &  4.169 $\pm$ 0.632  &  0.928 $\pm$ 0.012  &  0.812 $\pm$ 0.111  &  \bf{5.229 $\pm$ 1.721} & 0.108  \\ 
    
    Base learner 3D       &  0.809 $\pm$ 0.069  &  0.785 $\pm$ 0.235  & 4.121 $\pm$ 1.870 &  0.937 $\pm$ 0.008  & 0.799 $\pm$ 0.145  &  6.285 $\pm$ 3.108  &  0.13   \\ 

    Average ensemble  &  0.805 $\pm$ 0.073  &  0.708 $\pm$ 0.184   &  \bf{3.211 $\pm$ 0.923}  &  0.936 $\pm$ 0.011   &  0.752 $\pm$ 0.119  &   5.960 $\pm$ 2.526 & 0.2 \\ 
    
    \hline
    Our meta-learner \\ (only training data) &  0.823 $\pm$ 0.060  &   0.685 $\pm$ 0.164   &  3.224 $\pm$ 1.096  &  0.935 $\pm$ 0.010  &   0.763 $\pm$ 0.120  &  5.804 $\pm$ 2.670  &  0.21 \\ 
    
    Our meta-learner \\ (transductive)  & \bf{0.833 $\pm$ 0.054}  &  \bf{0.681 $\pm$ 0.178}   &  3.285 $\pm$ 1.370   &   \bf{0.939 $\pm$ 0.008}   &   \bf{0.733 $\pm$ 0.143}  &  5.670 $\pm$ 2.808  &  \bf{ 0.234 } \\ 
    
    \hline
\end{tabular}
\end{table*}

\section{Evaluation datasets and implementation details}

We evaluate our approach using two public datasets: (1) the HVSMR 2016 Challenge dataset \cite{pace2015interactive} and (2) the mouse piriform cortex dataset \cite{lee2015recursive}. 

\textbf{HVSMR 2016.}
The objective of the HVSMR 2016 Challenge \cite{pace2015interactive} is to segment the myocardium and great vessel (blood pool) in cardiovascular magnetic resonance (MR) images. 10 3D MR images and their corresponding ground truth annotation are provided by the challenge organizers as training data. The test data, consisting of another 10 3D MR images, are publicly available, yet their ground truths are kept secret for fair comparison.
The results are evaluated using three criteria: (1) Dice coefficient, (2) average surface distance (ADB), and (3) symmetric Hausdorff distance. Finally, a score $S$, computed as $S=\sum_{class}(\frac{1}{2}\mbox{\textit{Dice}} - \frac{1}{4}\mbox{\textit{ADB}} -\frac{1}{30}\mbox{\textit{Hausdorff}})$, is used to reflect the overall accuracy of the results and for ranking.

\textbf{Mouse piriform cortex.} Our approach is also evaluated on the mouse piriform cortex dataset \cite{lee2015recursive} for neuron boundary segmentation in serial section EM images.
This dataset contains 4 stacks of 3D EM images.
Following the previous practice \cite{lee2015recursive,neuron2}, the 2nd, 3rd, and 4th stacks are used for model training, and the 1st stack is used for testing. Also, as in \cite{lee2015recursive,neuron2}, the results are evaluated using the Rand F-score (the harmonic mean of the Rand merge score and the Rand split score).

\textbf{Implementation details.}
All our networks are implemented using TensorFlow \cite{abadi2016tensorflow}.
The weights of our 2D base-learners are initialized using the strategy in \cite{he2015delving}. The weights of our 3D base-learner and meta-learner are initialized with a Gaussian distribution ($\mu = 0$, $\sigma = 0.01$). All our networks are trained using Adam \cite{adam} with $\beta_1 = 0.9$, $\beta_2 = 0.999$, and $\epsilon = \mbox{1e-10}$. The initial learning rates are all set as $\mbox{5e-4}$. Our 2D base-learners reduce the learning rates to $\mbox{5e-5}$ after 10k iterations; our 3D base-learner and meta learner adopt the “poly” learning rate policy \cite{yu2017automatic} with the power variable equal to 0.9 and the max iteration number equal to 40k.
To leverage the limited training data, standard data augmentation techniques (i.e., random rotation with 90, 180, and 270 degrees, as well as image flipping along the axial planes) are employed to augment the training data.

For the HVSMR 2016 Challenge dataset, due to large intensity variance among different images, all the cardiac images are normalized to have zero mean and unit variance. We also employ spatial resampling to $1mm$ isotropically. For the mouse piriform cortex data, since the 3D EM images are highly anisotropic ($7\times7\times40 nm$), the 2D base-learners in the $xz$ and $yz$ views did not converge well. Thus, we only use the 3D base-learner and the 2D base-learner in the $xy$ view for this dataset.

\section{Experiments}

Because our meta-learner training does not require any manual-labeled data, our method can be easily adapted to the semi-supervised and transductive settings.
Thus, we experiment with the following three main settings to demonstrate the effectiveness of our method.
\begin{enumerate}
    \item To achieve fair comparison with the known state-of-the-art methods that cannot leverage unlabeled data, under the first setting, we train our meta-learner using only training data (the ``only training data'' entries in Tables \ref{tab:HVSMR_1} and \ref{tab:Neuron}).
    \item We show that our model can be improved under the semi-supervised setting in which we use additional unlabeled images to train our meta-learner (Table~\ref{tab:SS_heart}).
    \item  We show that improved results can be obtained under the tranductive setting in which we allow our meta-learner to utilize test data (the ``transductive'' entries in Tables \ref{tab:HVSMR_1} and \ref{tab:Neuron}). We emphasize that, although it might be less common to use test data for training in natural scene image segmentation, the transductive setting plays an important role in many biomedical image segmentation tasks (e.g., for making biomedical discoveries). For example, after biological experiments are finished, one may have all the raw images available and the sole remaining goal is to train a model to attain the best possible segmentation results for all the data to achieve accurate quantitative analysis.
\end{enumerate}

\subsection{Comparison with state-of-the-art methods when only using training data}

\noindent
\textbf{HVSMR 2016.}
Table~\ref{tab:HVSMR_1} shows a quantitative comparison with other methods in the leader board of the HVSMR 2016 Challenge \cite{pace2015interactive}.
Recall the two categories of the known deep learning based 3D segmentation methods (discussed in Section~\ref{Sec-Intro}). 
We choose at least one typical method from each category for comparison. (1) \cite{wolterink2016dilated} is based on the tri-planar scheme \cite{prasoon2013deep}, which utilizes \emph{three} 2D ConvNets on the orthogonal planes to predict a class label for each voxel. (2) VFN \cite{xia2018bridging} first trains \emph{three} 2D models with slices that are split from three orthogonal planes, respectively, and then applies a 3D ConvNet to fuse 2D results together. Besides, we compare our approach with state-of-the-art models (including 3D U-Net \cite{cciccek20163d}, VoxResNet \cite{chen2017voxresnet}, and DenseVoxNet \cite{yu2017automatic}).
Without using unlabeled data, our meta-learner outperforms these methods on nearly all the metrics and has a very high overall score, 0.215 (ours) \emph{vs.} $-0.161$ (DenseVoxNet), $-0.036$ (tri-planar), and $ 0.108 $ (VFN).

\noindent
\textbf{Mouse piriform cortex.}
Owning to the advanced components used in our base-learners (e.g., ResNet components \cite{he2016deep} and DenseNet components \cite{huang2017densely}), our 2D and 3D base-learners already achieve better results than the known state-of-the-art methods (Table \ref{tab:Neuron}). Nevertheless, from Table~\ref{tab:Neuron}, one can see that our meta-learner is able to (1) further improve the accuracy of the base-learners, and (2) achieve a result that is considerably better than the known state-of-the-art methods (0.9967 \emph{vs.} 0.9866).

\begin{table}
\centering
\caption{Quantitative results on the mouse piriform cortex dataset.}
\label{tab:Neuron}
\small
\begin{tabular}{cc}
  \hline
  {Method}&{$V^{Rand}_{Fscore}$}\\
  \hline
  N4 \cite{ciresan2012deep} & 0.9304 \\
  VD2D \cite{lee2015recursive} & 0.9463\\
  VD2D3D \cite{lee2015recursive} & 0.9720\\
  M$^2$FCN \cite{neuron2} & 0.9866\\
  Our 2D base-learner & 0.9948\\
  Our 3D base-learner & 0.9956\\
  Average ensemble of 2D and 3D & 0.9959\\
  Random-fit (only training data) & 0.9963\\
  NN-fit (only training data) & 0.9967\\
  Random-fit (transductive) & 0.9967\\
  NN-fit (transductive) & \bf{0.9970}\\
  \hline
\end{tabular}
\end{table}

\begin{table*}
\centering
\caption{Ablation experiments on the HVSMR 2016 dataset. ``GT'' represents ground truth and ``PL'' represents pseudo labels. Transductive learning setting: Test image data are involved as unlabeled data in model training.}
\label{tab:Ablation}
\setlength{\tabcolsep}{4.6pt}
    \begin{tabular}{cccccccccc} 
    \hline
    \multirow{ 2}{*}{ Setting } & \multicolumn{2}{c}{ Inputs }  &  \multirow{ 2}{*}{\shortstack{Supervision of \\ training set}}  &   \multirow{ 2}{*}{ \shortstack{ Transductive \\ learning} } &  \multirow{ 2}{*}{\shortstack{Supervision of \\ testing set}}   &  \multicolumn{2}{c}{ Training }  &  \multirow{ 2}{*}{ \shortstack{Overall \\ score} }\\
    \cline{2-3}  \cline{7-8} 
     & \shortstack{raw image ($x_i$)} & \shortstack{$S(PL_i)$}  &   &   &   & \shortstack{random fit}  & \shortstack{NN fit}  &   &  \\ 
    \hline

    S1  &           & \ding{51} & GT        &           &       &           &             &  0.075 \\ \hline
    S2  & \ding{51} & \ding{51} & GT        &           &       &           &             &  0.192 \\ \hline
    S3  & \ding{51} & \ding{51} & GT        & \ding{51} & PL    & \ding{51} &             &  0.217 \\ \hline
    S4  & \ding{51} & \ding{51} & GT + PL   & \ding{51} & PL    & \ding{51} &             &  0.205\\ \hline
    S5  & \ding{51} & \ding{51} & GT + PL   & \ding{51} & PL    & \ding{51} & \ding{51}   &  0.224 \\ \hline
    S6  & \ding{51} & \ding{51} & PL        &           &       & \ding{51} &             &  0.199  \\ \hline
    S7  & \ding{51} & \ding{51} & PL        &           &       & \ding{51} & \ding{51}   &  0.215 \\ \hline 
    S8  & \ding{51} & \ding{51} & PL        & \ding{51} & PL    & \ding{51} &             &  0.218 \\ \hline
    S9  & \ding{51} & \ding{51} & PL        & \ding{51} & PL    & \ding{51} & \ding{51}   &  \bf{0.234} \\ 
    \hline
\end{tabular}
\end{table*}

\subsection{Utilizing unlabeled data}
\noindent
\textbf{Semi-supervised setting.}
We conduct semi-supervised learning experiments on the HVSMR 2016 dataset. The training set of HVSMR 2016 is randomly divided into two groups evenly, $S_a$ and $S_b$. 
We conduct two sets of experiments. 
Under the setting of ``Group A", we first use $S_a$ to train base-learners using the original manual annotation; we then use $S_a \cup S_b$ to train our meta-learner with pseudo labels generated by the trained base-learners. For the overall training procedure, $S_a$ is labeled data and $S_b$ is unlabeled data. Testing phase utilizes the original test images in HVSMR 2016 dataset. The training \& testing procedures for ``Group B" follows the same protocol except that base-learners are trained with $S_b$. As shown in Table \ref{tab:SS_heart}, by leveraging unlabeled images, our approach can improve the model accuracy and generalize well to unseen test data. 

\noindent
\textbf{Transductive setting.}
In this setting, we use the full training data to train our base learners, and use the training and testing data to train our meta-learner. As discussed at the beginning of this section, the transductive setting is important for biomedical image segmentation applications and research. The ability to refine the model after seeing the raw test data (no annotation for test data) is another advantage of our framework. From Table \ref{tab:HVSMR_1} and Table \ref{tab:Neuron}, one can see that our meta-learner can achieve further improvements than using only the training data (0.234 \emph{vs.} 0.215 on the HVSMR dataset, and 0.9970 \emph{vs.} 0.9967 on the piriform dataset).

\subsection{Ablation study}
 
\noindent
\textbf{Average ensemble \emph{vs.} na\"ive meta-learner \emph{vs.} our best.} 
The results of the average ensemble of all the base-learners (the 2D and 3D models) are shown in Tables \ref{tab:HVSMR_1} and \ref{tab:Neuron}. One can see that the average ensemble is consistently worse than our meta-learner ensemble.
We also compare our meta-learner with the na\"ive meta-learner implementation (in which the outputs of the base-learners are used as input and the ground truths of the training set are used to train the meta-learner). Table~\ref{tab:Ablation} shows the results (the S1 row). One can see that the na\"ive meta-learner implementation is even worse than the average ensemble (probably due to over-fitting). This demonstrates the effectiveness of our meta-learner structure design and training strategy.

\noindent
\textbf{Random-fit + NN-fit \emph{vs.} Random-fit alone.}
Random-fit + NN-fit performs significantly better than Random-fit alone (Table~\ref{tab:Ablation}: S7$>$S6, S5$>$S4, S9$>$S8; Table~\ref{tab:Neuron}), which demonstrates that NN-fit can help the training procedure converge and thus improve the segmentation quality.

\noindent
\textbf{Model training using pseudo-labels \emph{vs.} ground truth.}
One may concern that our meta-learner training method totally discards manual-labeled ground truth even when it is available. This ablation study shows that our method can perform better without using any manual ground truth. We explore the following ways of utilizing ground truth. When using only the training data, we compare the difference between only ground truth (S2) and only pseudo-labels (S7). Table~\ref{tab:Ablation} shows that our training method can achieve better results (0.215 $>$ 0.192) when not using ground truth. When utilizing the test data (the transductive setting), we compare the difference between (1) only ground truth (S3), (2) mix of ground truth and psuedo-labels, i.e., using ground truth as the 5th version (S4 \& S5), and (3) only pseudo-labels (S8 \& S9). In Table~\ref{tab:Ablation}, one can see that (a) using pure ground truth or pure pseudo-labels achieves better results than mixing them together (probably due to the different nature of ground truth and pseudo-labels), and (b) using only pseudo-labels is still better than using ground truth (S8 $>$ S3). We think the reason that our method can work well with only pseudo-labels is because the pseudo-labels have already effectively distilled the knowledge from ground truth \cite{hinton2015distilling}.

\begin{table}
\centering
\caption{ Semi-supervised setting on HVSMR 2016 dataset.}
\label{tab:SS_heart}
\begin{tabular}{ccc}
  \hline
  {Group}& {Model} & {Overall score}\\
  \hline
  \multirow{2}{*}{ A } &  Base-learner 3D & -0.036  \\
   & Meta-learner &  $ 0.063 $\\  \hline
   
   \multirow{2}{*}{ B } &  Base-learner 3D &  -0.045\\
   & Meta-learner & 0.038  \\  \hline
\end{tabular}
\end{table}

\section{Conclusions}
In this paper, we presented a new {\em ensemble learning} framework for 3D biomedical image segmentation that can retain and combine the merits of 2D and 3D models. Our approach consists of (1) diverse and accurate base-learners by leveraging diverse geometric and model-architecture perspectives of multiple 2D and 3D models, (2) a fully convolutional network (FCN) based meta-learner that is capable of learning robust visual features/representations to improve the base-learners' results, and (3) a new meta-learner training method that can minimize the risk of over-fitting and utilize unlabeled data to improve performance. Extensive experiments on two public datasets show that our approach can achieve superior performance over the state-of-the-art methods.

\section{Acknowledgments}

This research was supported in part by the U.S. National Science Foundation through grants CCF-1617735, IIS-1455886, and CNS-1629914.

\bibliographystyle{aaai}
\bibliography{reference}

\end{document}